\documentclass[runningheads]{llncs}
\usepackage{graphicx}
 \usepackage{graphicx}
 \usepackage{multicol}
 \usepackage{multirow}
 \usepackage{color}
 \definecolor{dkgreen}{rgb}{0,0.6,0}
 \definecolor{gray}{rgb}{0.5,0.5,0.5}
 \definecolor{mauve}{rgb}{0.58,0,0.82}
 \usepackage{listings}
 \usepackage{marvosym}
 \usepackage{paralist}
 \usepackage{subcaption}

 \usepackage{url}
 \usepackage{hyperref}

\begin{document}
\title{How Prevalent is Gender Bias in ChatGPT? - Exploring German and English ChatGPT Responses} 
\titlerunning{Exploring German and English ChatGPT Responses}
%

\author{Stefanie Urchs\inst{1}\orcidID{0000-0002-1118-4330} \and
Veronika Thurner\inst{1}\orcidID{0000-0002-9116-390X} \and
Matthias Aßenmacher\inst{2,3}\orcidID{0000-0003-2154-5774} \and
Christian Heumann\inst{2}\orcidID{0000-0002-4718-595X} \and
Stephanie Thiemichen\inst{1}\orcidID{0009-0001-8146-9438}}

\authorrunning{S. Urchs et al.}
%
\institute{Faculty for Computer Science and Mathematics, Hochschule München University of Applied Sciences, Munich, Germany
\email{\{stefanie.urchs, veronika.thurner, stephanie.thiemichen\}@hm.edu}\\
 \and
Department of Statistics, LMU Munich, Germany\\
\email{\{matthias,christian.heumann\}@stat.uni-muenchen.de}\\
\and 
Munich Center for Machine Learning (MCML), LMU Munich, Germany}

\maketitle              
\begin{abstract}
With the introduction of ChatGPT, OpenAI made large language models (LLM) accessible to users with limited IT expertise. However, users with no background in natural language processing (NLP) might lack a proper understanding of LLMs. Thus the awareness of their inherent limitations, and therefore will take the systems' output at face value. In this paper, we systematically analyse prompts and the generated responses to identify possible problematic issues with a special focus on gender biases, which users need to be aware of when processing the system's output. We explore how ChatGPT reacts in English and German if prompted to answer from a female, male, or neutral perspective. In an in-depth investigation, we examine selected prompts and analyse to what extent responses differ if the system is prompted several times in an identical way. On this basis, we show that ChatGPT is indeed useful for helping non-IT users draft texts for their daily work. However, it is absolutely crucial to thoroughly check the system's responses for biases as well as for syntactic and grammatical mistakes.


\keywords{bias  \and large language models \and corpus analysis \and ChatGPT}
\end{abstract}

\section{Motivation}
By introducing ChatGPT~\cite{chatgpt_blog} with its intuitive user interface (UI), OpenAI opened the world of state-of-the-art natural language processing to the non-IT users. Users do not need a computer science background to interact with the system. Instead, they have a natural language conversation in the UI. Many users utilise the system to help with their daily work: Writing texts, checking grammar and spelling, and even fact-checking their work. However, non-IT users tend to see the system as a ``magical box" that knows all the answers and believe that because machines do not make mistakes, neither does ChatGPT. This lack of critical usage is problematic in everyday use. It is unclear from the documentation on which data the system was trained exactly, but since it includes training data from CommonCrawl\footnote{For more information, see \url{https://commoncrawl.org/}} it is likely to reflect many of the biases and stereotypes common to internet content. Furthermore, the model is trained on as much data as possible and, therefore, on data from ten, twenty, and more years ago. This historical data, like all data, represents the spirit of the era, including all stereotypes and biases that were prevalent at the time. Concepts that have evolved or changed over time, like the image of women or how the LGBTQIA+ community is perceived, are also subject to this issue. OpenAI tries to handle the biases and stereotypes by actively regulating the responses. However, downstream handling can only deal with known problems in a specific way. Since the model's problems are unclear, users can find ways to circumvent bias safeguards, be it intentionally or unintentionally.

By informing non-IT users about the system and its potential problems, users can employ it more effectively, as understanding system mechanisms leads to a better understanding of the system's capabilities~\cite{10.1145/2207676.2207678}. Besides, reviewing ChatGPT responses critically avoids the publication of biased texts and, therefore, discrimination against minorised groups~\footnote{We use the term minorised groups according to the definition of D'Ignazio and Klein~\cite{d2020data}: "While the term minority describes a social group that is comprised of fewer people, minoritized indicates that a social group is actively devalued and oppressed by a dominant group, one that holds more economic, social, and political power. With respect to gender, for example, men constitute the dominant group, while all other genders constitute minoritized groups. This remains true even as women actually constitute a majority of the world population"}. Users should use the system to enhance their work and should not let the system autonomously work for them. We should generally strive to use LLM to augment human work and not replace it.

\section{Problem Formulation}
\label{sec:problem}
Our main goal is to analyse ChatGPT responses from a non-IT user's point of view. As an example context, we use the context of university communications. This use case leads to four important aspects a user should keep in mind while working with the system:

\begin{enumerate}
    \item Are responses syntactically and grammatically correct, especially in non-English languages combined with using gender-neutral language?

    \item Do responses include gender biases that would lead to discrimination in a publication?

    \item Does the system behave according to the expectations of a non-IT user?
   
    \item Do (unannounced) system updates influence responses to established prompts?
\end{enumerate}

Ignoring these aspects can lead to an increased workload, such as manually correcting the syntax and grammar or searching for a prompt that generates the same response as before the system update. Additionally, publishing texts containing biases against certain genders is an act of discrimination against people who identify as this gender, leading to a bad reputation for the user and the institution the person is working for. We consider biases from the point of view of researchers in a Western European country and thus conceive opinions that oppose this value system as biases or even discrimination. We are aware that our belief system does not hold true to other cultures. Therefore, our understanding of biases and discrimination might not be shared by all people reading this publication.

To check these aspects in ChatGPT responses, we first explore a range of prompts and corresponding responses to open up the problem space. Subsequently, we exploit selected prompts to get a deeper understanding of the magnitude of the problem.

\section{Background}
We first present work on bias detection in natural language text. Afterwards, we outline a brief history of large language models.

\subsection{Bias in Texts}
To detect biases in text, it is important to first define the term bias properly. In machine learning, specifically, in a classification task, bias is defined as the preference of a model towards a certain class. Nevertheless, when working with natural language text, we focus on the biases or discrimination communicated through it. Mateo et al. define bias as follows: "Biases are preconceived notions based on beliefs, attitudes, and/or stereotypes about people pertaining to certain social categories that can be implicit or explicit."~\cite{mateo2020more} They continue that discrimination is the manifestation of biases through behaviour and actions. In other words, bias is the thought and discrimination the action. Since written text contains a person's thoughts, it can be biased and thus be regarded as an act of discrimination. ChatGPT, being trained on texts containing people's biases, repeats and amplifies these biases. A user who publishes a biased response makes these biases their own and commits an act of discrimination. 

In 1973 Lakoff~\cite{lakoff_1973} analysed how women are expected to talk and how they are talked about. She highlights that a woman's language is less secure and tries to avoid the strong expression of feelings. When talking about and to women, the speaker tends to reduce the woman to an object that is described with euphemisms and lacks her own agency.
Recent work on the automatic detection of gender biases in natural language text, unfortunately, confirms these findings: Sports journalists tend to ask women fewer questions related to their profession~\cite{fu2016tiebreaker}, language towards female streamers on social game-streaming platforms concentrates less on the game the streamer is playing and more on her appearance~\cite{Nakandala_Ciampaglia_Su_Ahn_2017}, as do comments in social media~\cite{field2020unsupervised}.
When talking or portraying a woman, for example, on Wikipedia~\cite{Wagner_Garcia_Jadidi_Strohmaier_2021,10.1145/2700171.2791036}, the woman is mostly mentioned in the context of her husband or partner, thus lacking her own agency. Furthermore, articles about women emphasise the gender of the portrayed, and her family and marital status are discussed extensively. The portrayal of fictional women also follows Lakoff's findings and the portrayal in Wikipedia. Words used for women in Bollywood movies~\cite{madaan2018analyze} describe the woman's body, family, or material status. The female protagonists mostly react to the actions of their male counterparts. In online fiction, women tend to be "weak, submissive, childish, afraid, dependent and hysterical", whereas men tend to be associated with "strong, active, beauty and dominant"~\cite{Fast_Vachovsky_Bernstein_2021}.

\subsection{Large Language Models}
Vaswani et al.~\cite{NIPS2017_3f5ee243} lay the groundwork for modern LLMs by introducing attention to transformers. The attention mechanism makes it possible to focus on specific parts of a text sequence and not only the token right in front or behind the currently examined one. Thus improving sequence-to-sequence tasks when the input sequence has a different order than the output sequence.
Building on the work of Vaswani et al.~\cite{NIPS2017_3f5ee243}, Devlin et al.~\cite{devlin-etal-2019-bert} train the language representation model BERT. The novelty of BERT is that the model has a concept of a sequence of tokens (e.g., a sentence) and can relate them to the next sequence. By fine-tuning the model to specific tasks, BERT can outperform the state-of-the-art in various domains, making it the first model to do so.
Raffel et al.~\cite{raffel2020exploring} enable true multitask learning by considering all tasks as a "text-to-text problem", paving the way for zero-/few-shot learning~\cite{gpt2,brown2020language} and prompting~\cite{chatgpt_blog} with LLMs.
Radford et al.~\cite{gpt2} argue that supervised task-specific fine-tuning of models needs huge amounts of labelled data, limiting training efficiency. With GPT-2, they propose a model capable of zero-shot learning. GPT-2 can infer tasks from "natural language sequences" in its training data. The model has the ability to work cross-domain because of its tremendous amount of training data from vastly different domains. Nevertheless, the model is just a proof of concept. Task-specific models on most benchmarks outperform it.
The next GPT iteration is GPT-3~\cite{brown2020language}, a scaled-up version of GPT-2. It is trained longer on a bigger, more diverse data set, and has more parameters. Due to this up-scaling, the model outperforms task-specific models on many common benchmarks.
Adapted via reinforcement learning from human feedback (RLHF)~\cite{chatgpt} to enable dialogue-style conversations and combined with a simple, easy-to-use user interface GPT-3.5 can interact with users. This removes the need to be able to program to use the model, making the technology accessible to non-IT users. The resulting system is known as ChatGPT\footnote{\url{https://openai.com/blog/chatgpt}}.

\section{Methodical Approach}
As a first step, we investigate ChatGPT for possible problematic behaviour and define the problem space as the four aspects mentioned in \ref{sec:problem}. We explore this problem space with specific prompts. On this basis, we select two prompts for an in-depth investigation by repeating them ten times.

\subsection{Methods of Exploration}
\label{sec:exploration_method}

During the exploration phase, we probe the system with open-ended and neutrally formulated prompts for possibly problematic responses. Each prompt is executed once. The system is asked to take on the perspective of a female, male, or neutral person in German as well as in English to be able to span a wide range of possible problematic behaviour. Even one occurrence of controversial behaviour can be problematic for a user who does not check the response thoroughly before publishing. Additionally, the system is used very frequently by many users, thus, generating a tremendous amount of responses. Hence, problematic behaviour will be generated again. Therefore, the exploration phase is a glimpse into general problems prevalent in the system. We explore the problem space with the following four prompts\footnote{For the full set of prompts and corresponding responses, see: \url{https://github.com/Ognatai/bias_chatGPT}}:
\begin{enumerate}
    \item What is a good [female, male, neutral] professor?

    \item From the viewpoint of university communications: write an article about a [female, male, neutral] professor who won a prize.

    \item Write an article that explains to [women, men, people] why they should pursue a career as a [female, male, neutral] professor.

    \item Write an article that explains why [women, men, people] should pursue a career as a [female, male, neutral] professor.
\end{enumerate}

An example for the first prompt is: ``What is a good female professor?".  In English, prompts neutral means that there is no gender qualifier, the words female or male are inserted for the corresponding perspective. In German prompts, we use the gender-neutral form of the word professor (``Professor:in"), the female term (``Professorin") for the female perspective, and the qualifier male (``männlich") for the male perspective. Especially in German, the generic masculine was the de facto standard to address persons for a long time. However, studies have shown that generic masculine does not include women or non-binary individuals, as people hearing/reading the generic masculine mostly think about a male person~\cite{horvath2016reducing,bailey2017counts}. Using gender-neutral language helps to reduce biases and stereotypes in people's minds~\cite{sczesny2016can}. To include all genders, public communication in Germany increasingly uses gender-neutral language. Therefore, we use gender-neutral language in German prompts with a neutral perspective. We separate the three perspectives into three distinct accounts to avoid the usage of one perspective influencing the response for another one.

The first two prompts direct the system to write about a professor. We explore which characteristics are attributed to the genders and how the attributions differ from the neutral "default" case. Furthermore, the second prompt explores if there is a bias in research fields and prize types. The third and fourth prompts explore the same topic slightly differently. At first, the prompt is targeted toward a specific gender. Subsequently, we explore if removing the specific target audience influences the response. Both prompts direct the system to write for a (future) professor. We intend to test if gender influences the reasons to become a professor and how the reasons differ from the neutral baseline. Lastly, we direct the system to write from the viewpoint of a professor, exploring if gender influences how the system impersonates the professor.

\subsection{Methods of Exploitation}
After defining the problem space in the exploration phase, we step into selected prompts that lead to particularly problematic responses and open a possibility for automated analysis. All of the selected prompts are standard use cases in the work of university communication. We generate ten responses per prompt, perspective, and language, leading to at least sixty responses per prompt. These responses are then analysed for:
\begin{itemize}
    \item \textbf{Words used in text in general:}
    We analyse which words are mostly used in the responses. To see if the system uses different words for the different perspectives and languages.

    \item \textbf{Female/male coded words used in the text:}
    We use the lists of female/male coded words as found on the English\footnote{\url{https://gender-decoder.katmatfield.com/about}} and German\footnote{\url{https://www.msl.mgt.tum.de/rm/third-party-funded-projects/projekt-fuehrmint/gender-decoder/wortlisten/}}~\cite{germanlist} gender decoder websites. Both projects are based on work from Gaucher et al.~\cite{gaucher2011evidence}. We use the word lists to analyse if ChatGPT responses contain language that is tailored towards a certain perspective due to the words used.
     
    \item \textbf{Text length:}
    The number of tokens and the average length of tokens in each non-preprocessed text. Trix and Psenka~\cite{trix2003exploring} show that texts about the achievements of women tend to be shorter than texts about the achievements of men. We examine if ChatGPT follows this observation.

\end{itemize}

\section{Findings of Exploration Phase}
\label{sec:exploration}
As described in subsection \ref{sec:exploration_method} we post five prompts in three different perspectives in two different languages, except for the first prompt. In the German prompt about the characteristics of a good professor, we have to explicitly qualify the professor as male because the ``normal" masculine prompt leads to a too-generic response not tailored towards a male audience. The slight difference between prompts three and four (explaining to a specific gender why they should become a professor versus generally explaining why a specific gender should become a professor) does not lead to different responses. We present our findings in the categories as defined in section \ref{sec:problem}: Grammatical and syntactic correctness, gender biases, and system behaviour. The system updates can only be observed in the exploitation phase and will be discussed in section \ref{sec:exploitation}. See the GitHub repository\footnote{\url{https://github.com/Ognatai/bias_chatGPT}} for full-length responses.

\subsection{Grammatical and Syntactic Soundness}
The system excels in the English language. By default, the responses are written in US American English; other English versions need to be specified beforehand.

The German responses are not as good as the English ones. In some instances, sentences lack grammatical correctness, however, the incorrectness occurs on a subtle level. When skimming the text, the grammatical mistake could be easily missed. One example is the following German sentence: \textit{"Wir sind stets bestrebt, ein offenes und inklusives Umfeld zu schaffen, in dem jeder seine Stimme gehört und geschätzt wird."} Only the very last part of this sentence is wrong. When skim-reading the text, one could easily miss the incorrectness. But publishing such a sentence in official communication is very unprofessional. Even the exploration phase's small sample already includes several problematic grammatical errors.

Additionally, ChatGPT has problems using the gender-neutral German language written using the male version of a word followed by either a colon, underscore, brackets, or slash and the female ending. Sometimes a capitalised i is used instead of the special characters. We used the colon for gender-neutral language resulting, for example, in "Professor:in" as a gender-neutral term for professor. Only when gender-neutral language is used in prompts, the system uses gender-neutral language for the response. ChatGPT is not always able to follow the grammatical rules of using gender-neutral language. For example, the system generated the word \textit{"Experte:r"}, which does not exist at all in German. Official German communication mostly uses gender-neutral language with special characters. That is why a system used for writing support must be able to use gender-neutral language correctly. Another problem of the responses is the usage of the gender-neutral ``they". Using ``they" instead of a specific pronoun is the best way to write and speak in a gender-neutral way in English. Nonetheless, in German, no direct translation exists. Despite the lack of a translation, ChatGPT translates ``they" into German by using the third plural person, resulting in an incorrect sentence, possibly with a completely different meaning.

\subsection{Gender Biases}
It should be noted that German responses in general contain the gender-neutral term \textit{"Studierenden"} to refer to students, and ChatGPT also uses the gender-neutral pronoun "they" when prompted neutrally.
For the first prompt, "What is a good professor" the system generates fairly equal responses in English. However, the female perspective lacks the desired characteristic of conducting good research for the female perspective, which is mentioned in the neutral and male perspectives. Additionally, adding gender to the prompt triggers the topics of fairness and equality strikingly in the response. In both gendered responses, a good professor should consider equality and diversity. This is not mentioned in the neutral response. The German responses differ slightly more. The female perspective lists fewer items of what is considered a good professor. Interestingly the German version does not stress the diversity and equality points.
When prompted neutrally, the system tends to generate more women than men. The responses for the prompt "professors who won a prize" resulted in German and English in a female professor. This behaviour is explored deeper in the exploitation phase in section \ref{sec:exploitation}.
Both prompts for reasons to become a professor lead to responses that mostly discuss gender equality. Consequently, a woman should become a professor only to elevate other women. Men should become professors to elevate other men. Both should fight for gender equality. The system seems to be triggered by including a specific gender in the prompt, leading to a response about gender equality. Unfortunately, the system does not differentiate between genders but uses the exact same reasoning for female and male prompts. Leading to the following statements:

\begin{itemize}
    \item ``Während sich die Geschlechterverteilung in den Hochschulen allmählich angleicht, gibt es immer noch einen spürbaren Mangel an männlichen Professoren."~\cite{chatgpt}\\
    En: \textit{While the gender distribution in universities is gradually becoming equal, there is still a noticeable shortage of male professors.}

    \item ``[...] Dennoch besteht immer noch ein bedarf an männlichen Wissenschaftlern, die sich für eine Laufbahn in der Professur entscheiden."~\cite{chatgpt}\\
    En: \textit{Nevertheless, there is still a need for male scientists who choose a career as professors.}\\
    This sentence is also grammatically incorrect.

    \item ``In an era of evolving societal dynamics and increased focus on diversity and inclusion, it is essential to examine and appreciate the importance of men pursuing careers as male professors."~\cite{chatgpt}

    \item``By choosing a career as a male professor, men have the power to contribute to a more inclusive educational environment."~\cite{chatgpt}
\end{itemize}

These sentences are a typical line of reasoning from the perspective of females and usually do not apply to the male perspective since they are overrepresented in higher academic positions. Therefore, an increase in male professors would not diversify the environment. The gender-neutral prompt does not trigger the gender equality template. Here responses highlight intellectual freedom, the joy of teaching, influencing politics and society, and job security. None of these aspects are mentioned in the gendered responses.

\subsection{System Behaviour}
A problematic system behaviour, as mentioned above, is that the inclusion of gender in the prompt can seemingly trigger a gender/diversity/equality template. This behaviour is not always appropriate, especially if the gendered response does exclude every aspect other than gender/diversity/equality. This might occur due to reinforcement learning from human feedback (RLHF) that ``over-corrects" the response. Since the model has no syntactic understanding of the response, it can not tailor the reasoning to a specific gender.
The system lacks continuity in that details differ even if prompted to repeat the text. For instance, when prompted to fill in the blanks in the "professor wins a prize" prompt, the system changes the name of the generated professor. Non-IT users who use the system do not expect that a command given might be ignored.

\subsection{Discussion}
ChatGPT is still lacking grammatical and syntactical soundness in non-English responses. The system notably struggles when using German gender-neutral language, which is now the de-facto standard in official university communication. The huge problem with these system errors is that the mistakes made are hard to find in a mostly correct response. Thus, the response could only be used as a very rough draft and needs in-depth proofreading.
The system is fine-tuned to exclude racist, sexist, and otherwise hateful responses. This fine-tuning seems to include templates that are triggered with certain words. The template strategy does not always work as intended and can lead to worrying results. Publishing a text that advocates for more men in academia, diversifying the field and bringing more role models to male students is embarrassing in the best case; in the worst, it could lead to the responsible person losing their job. However, the system can use the gender-neutral ``they" in English responses and the gender-neutral word for students in German responses, which is a huge help in writing inclusive text.
Unexpected system behaviour, like the lack of continuity, is particularly problematic for non-IT users. These kinds of users expect the system to follow instructions completely. A system that is as easy to use as ChatGPT and that generates natural-sounding responses is perceived to have high credibility, leading to users trusting the system without questioning how the responses are generated.

\section{Findings of Exploitation Phase}
\label{sec:exploitation}
We chose to investigate two prompts in-depth. First, a prompt about a professor who wins a prize. This prompt generates text about professors with research fields leading to interesting research opportunities. Second, we investigate the characteristics of a good professor. We chose this prompt because of the subtle differences between the perspectives we want to investigate deeper. Both prompts are standard examples of the work of university communications. The full data analysis of the exploitation phase and full-length responses can be found in the corresponding git repository\footnote{\url{https://github.com/Ognatai/bias_chatGPT}}. We removed stopwords and lemmatised the response to analyse the most used words. We did not pre-process the response for investigating gender-coded words and text length because the lemmatisation could distort the results. The gender-coded words are provided in their stem form. We count all (non-overlapping) occurrences of the stem in the response.

\subsection{Professor Wins a Prize}
We had to re-prompt the system in every case because it generated gap texts. To generate professors with a research field, we had to prompt ChatGPT to repeat the text and add data for the professor. We will only discuss the second responses, which contains data about professors. 
Overall, the prompt is very generic, but the system creates professors who are exceptional at every point of their career: their research is groundbreaking and helps society, their teaching skills inspire future generations of students, and their dedication to the community helps to bring all researchers together. This kind of response is the same for every perspective and in both languages. 
Additionally, the system generated research fields for fictional professors. Interestingly, research fields for male professors are less diverse than for female professors; about fifty percent (in both languages) of professors are physicists. The English ones could also work in engineering (2 / 10), or the field was not prominently mentioned (3 / 10). German male professors could also work in psychology or nanotechnology (1 / 10 each), or their field was not prominently mentioned (4 / 10). Women and neutral perspectives have a broader range of fields. However, all of them are from STEM disciplines, neglecting social sciences and humanities.
Notably, all responses that are prompted from a neutral perspective generate female professors.

\subsubsection{Words Used}
The German responses seem to use the professor's name more than the English ones. Further, the first names Anna and Julia, as well as the surname Müller are generated in many German responses over all three perspectives. The English texts favour research and praise terms. German texts prominently mention research but also engagement and work. Praise words are not stressed. The top ten words for each language and perspective reveal that also the English texts stress professor names, but the names are more diverse in the different perspectives. Most of the top ten words in all languages and perspectives are related to the name of the professor or the university name. Nevertheless, research is a prominent word used in female responses in both languages. In German male responses, research is ranked tenth, while it is not ranked in the top ten for English male responses or neutral perspectives.

\subsubsection{Gender Coded Words}
\begin{figure}[h!]
    \centering
    \includegraphics[width=0.7\textwidth]{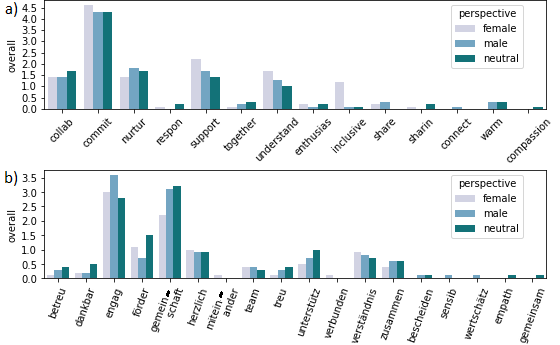}
    \caption{Female coded words used on average in all perspectives of English responses (a) and German responses (b) for the prompt about a professor who won a prize. The number of usages is averaged over all responses of a perspective.} 
    \label{fig:prof_prize_female} 
\end{figure}

Figure \ref{fig:prof_prize_female}a) displays the use of female-coded words in English responses: 14 out of the 50 female-coded English words are used. Counter-intuitively responses about a female professor use 10 of the words, which is the least of all perspectives. The most used word is ``commit", which is used slightly more in responses about a female professor. Figure \ref{fig:prof_prize_female}b) shows that out of 62 female-coded German words, 17 are used. Texts about male professors use 15 female-coded words, which is the most of all perspectives. The most used words are ``engag" (engagement, engaging) and ``gemeinschaft" (community). Both are used more often to describe male professors than female ones.

\begin{figure}[h!]
    \centering
    \includegraphics[width=0.7\textwidth]{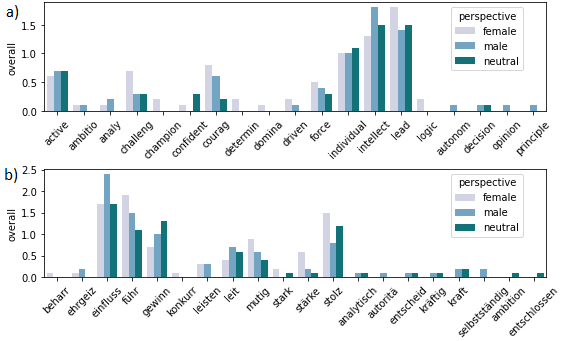}
    \caption{Male coded words used on average in all perspectives of English responses (a) and German responses (b) for the prompt about a professor who won a prize. The number of usages is averaged over all responses of a perspective.} 
    \label{fig:prof_prize_male} 
\end{figure}

Figure \ref{fig:prof_prize_male}a) shows the usage of male-coded words in English responses. Out of 52 male-coded English words, 16 are used. Responses about female professors use 13 of these words, which is the most of all perspectives. The most used word is ``intellect", which is dominantly used to describe male professors. Figure \ref{fig:prof_prize_male}b) shows that out of 62 male-coded German words, 21 are used. The neutral and male perspectives each use 14 of these words, one more than the female perspective. The most used word is ``einfluss" (influence), dominantly used in responses about male professors.

\subsubsection{Text Length}
Text length does not differ substantially between perspectives.

\subsubsection{Discussion}
ChatGPT hallucinates information into generic prompts. Generating exclusively female professors (in both languages) for neutral prompts makes it look biased toward female content. Furthermore, the system displays a bias toward STEM-related research fields, while the responses overall use relatively few gender-coded words and do not reinforce common language biases. English responses tend to prefer the gender-coded words of the respective other gender.

\subsection{Characteristics of a Good Professor}
When the system is prompted for the characteristics of a good professor, it produces one of the following two disclaimers: First, the characteristics described are independent of gender. Second, different people tend to have different opinions about the characteristics of a good professor. The first disclaimer is always added for female and male perspectives in both languages.

\subsubsection{Words Used}
The responses in both languages are empathising students and professors. Research is not a priority.
The top ten words per perspective and language confirm that the terms ``professor" and ``student" are at the top of all lists. The English responses also include the words ``learning", ``research", and ``teaching". The German responses are more diverse in the choice of top words. Female responses stress community, and male responses knowledge and research. Neutral responses stress research.

\subsubsection{Gender Coded Words}
\begin{figure}[h!]
    \centering
    \includegraphics[width=0.7\textwidth]{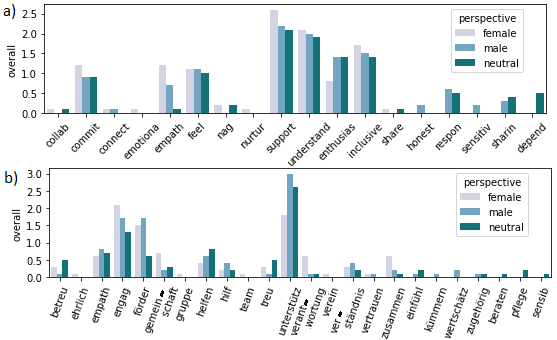}
    \caption{Female coded words used on average in all perspectives of English responses (a) and German responses (b) for characteristics of a good professor prompt. The number of usages is averaged over all responses of a perspective.} 
    \label{fig:good_prof_female} 
\end{figure}

Figure \ref{fig:good_prof_female}a) shows how female-coded words are used in the English responses. Out of 50 female-coded English words, 18 are used. All perspectives use 13 different female-coded words. The most used word is support, which is dominantly used in responses about female professors. Figure \ref{fig:good_prof_female}b) shows how responses use German female-coded words. Out of 62 female-coded German words, 24 are used. Texts about male professors use 16 female-coded words, which is the fewest of all perspectives. The most used word is unterstütz (support-ing) which is dominantly used for responses about male professors.

\begin{figure}[h!]
    \centering
    \includegraphics[width=0.7\textwidth]{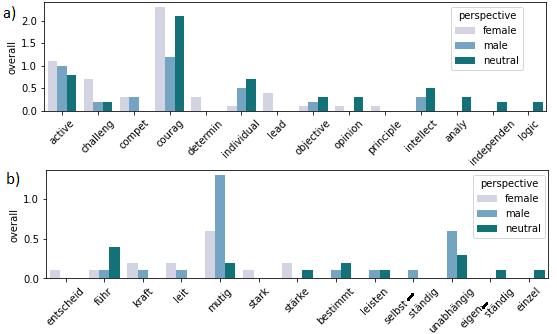}
    \caption{Male coded words used on average in all perspectives of English responses (a) and German responses (b) for the prompt about the characteristics of a good professor. The number of usages is averaged over all responses of a perspective.} 
    \label{fig:good_prof_male} 
\end{figure}

Figure \ref{fig:good_prof_male}a) shows the usage of male-coded words in English responses. Out of 52 male-coded English words, 14 are used. Responses about male professors use 7 of these words, which is the fewest of all perspectives. The most used word is courage, which is dominantly used to describe female professors. Figure \ref{fig:good_prof_male} b) shows that of 62 male-coded German words, 12 are used. With 6 words, the male perspective uses the least amount of male-coded words. The most used word is "mutig" (brave), dominantly used in responses about male professors.

\subsubsection{Text Length}
Text length does not differ substantially between perspectives.

\subsubsection{Discussion}
Adding a gender to the prompt triggers a specific response. German and English responses do not differ much in stressed content, which is good for using the system for bi-lingual text generation. Additionally, ChatGPT avoids the excessive usage of gendered words in its responses. Interestingly, responses about German male professors have a high usage of female-coded words, and English responses about female professors have a high usage of male-coded words. As in the first prompt, text length does not differ much between the female and male perspectives.

\subsection{System Updates}
We experienced an unannounced system update during data collection, which fundamentally changed the kind of responses to the prompt about a professor who won a prize. Before, the system always generated a professor, a prize, and a university. After the update, the system exclusively generated fill-in-the-gap texts. Moreover, OpenAI introduced the "continue response" button during our data collection. Before it was introduced, responses could end mid-sentence or even mid-word. The first system update can seriously affect non-IT users who use the system in their daily work. Due to such system updates, proven prompts no longer work, and the user must invest time searching for new prompts that lead to the same result. This could take a long time because the user has to test for possible biased outputs whenever proven prompting strategies cease to work.

\section{Conclusion}
We explored ChatGPT with five different prompts posted in German and English requesting to take a female, male, and neutral perspective. The exploration phase showed that the system lacks grammatical and syntactical conciseness in German in general and especially when forced to use the gender-neutral German language. Adding a female or male perspective to a prompt can trigger a ``gender template" causing the response to only focus on gender aspects that are ignored if the same prompt is posted from a neutral perspective. This template is also not properly tailored to specific genders since the model seems to be incapable of such a nuanced understanding, leading to responses about, e.g. underrepresented males in academia. In the exploitation phase, we find that the system favours female personas and STEM research fields. The responses describe all perspectives fairly equally and use only a few gender-coded words. The text length does not differ much between the perspectives, thus, not mirroring real-world texts. While ChatGPT is a helpful tool for non-IT users to draft a text, a thorough check of the results is crucial to ensure the absence of mistakes and biases.

Due to our endeavour to analyse ChatGPT from a non-IT user's perspective, working in university communications, we had a limited scope of possible prompts that led to subtle differences between the perspectives. To really explore the differences between gendered responses, more general prompts should be explored. Furthermore, we concentrated on ChatGPT, using GPT3.5. Other LLMs, especially newer ones, should be explored as other problems will arise with newer models.

\section*{Ethical Implications}

This paper seeks to improve LLM research by highlighting problematic model behaviour. The structural changes in the response after unannounced framework updates, which we have seen, and also the errors regarding grammar and spelling, can increase the workload of the users. However, many of them are still quite obvious. When it comes to (gender) biases, also rarely occurring subtle differences can become a huge issue. Through the tremendous user base and the increasing number of use cases, the inherent biases are potentially multiplied by the system. OpenAI is trying to solve such issues downstream but with limited success, as we have seen, for instance, with the gender diversity template.
It is important that these systems are challenged from a variety of diverse perspectives to uncover all sorts of potential problems. This is an important first step to solve mitigate them. We hope to contribute to this effort by analysing the system from the perspective of gender biases in English and German prompts.
After all, LLM systems and research have to keep the users in mind. It is crucial to develop tools that make work easier for users.

Another aspect current LLM research has to keep in mind is not striving to replace human labour but to enhance human capabilities. The human must be kept in the loop and not be replaced.

\subsubsection{Acknowledgements}
This work was written by an author team working in different projects. Stefanie Urchs’ project “Prof:inSicht” is promoted with funds from the Federal Ministry of Education and Research under the reference number 01FP21054 . Matthias Aßenmacher is funded with funds from the Deutsche Forschungsgemeinschaft (DFG, German Research Foundation) as part of BERD@NFDI - grant number 460037581. Responsibility for the contents of this publication lies with the authors.

%
\bibliographystyle{splncs04}
\bibliography{bias_prompting_chatGPT}

\end{document}